%% file: main.tex
\documentclass[10pt,twocolumn,letterpaper]{article}
\usepackage{times,graphicx,amsmath,amssymb,booktabs,array,tabularx,xcolor}
\usepackage[numbers,sort&compress]{natbib}
\usepackage[hyphens]{url}
\usepackage[format=plain,labelformat=simple,labelsep=period,font=small]{caption}
\usepackage{microtype}
\definecolor{refblue}{rgb}{0.21,0.49,0.74}
\usepackage[pagebackref,breaklinks,colorlinks,allcolors=refblue,hypertexnames=false]{hyperref}

\makeatletter
\font\subheadingfont=ptmb scaled 1100
\newcommand{\sectionheading}{\@startsection{section}{1}{\z@}
  {-10pt plus -2pt minus -2pt}{7pt}{\large\bfseries}}
\newcommand{\subsectionheading}{\@startsection{subsection}{2}{\z@}
  {-8pt plus -2pt minus -2pt}{5pt}{\subheadingfont}}
\newcommand{\numberedheading}[2]{#1{\texorpdfstring{\hskip-1em.~}{}#2}}
\renewcommand{\section}{\@ifstar{\sectionheading*}{\numberedheading\sectionheading}}
\renewcommand{\subsection}{\@ifstar{\subsectionheading*}{\numberedheading\subsectionheading}}
\renewcommand{\@maketitle}{%
  \newpage\null\vskip .375in
  \begin{center}
    {\Large\bfseries\@title\par}
    \vspace*{24pt}%
    {\large\lineskip .5em
     \begin{tabular}[t]{c}
       \raisebox{0pt}[0pt][0pt]{\begin{tabular}{c}\@author\end{tabular}}
     \end{tabular}\par}
    \vskip .5em\vspace*{12pt}
  \end{center}}
\makeatother
\renewenvironment{abstract}
  {\centerline{\large\bfseries Abstract}\vspace*{12pt}\noindent\itshape\ignorespaces}
  {\vspace*{12pt}}

\newcolumntype{L}{>{\raggedright\arraybackslash}X}
\newcolumntype{C}{>{\centering\arraybackslash}X}
\newcommand{\dicesummary}[2]{\mbox{#1{\scriptsize\,$\pm$\,#2}}}
\newif\ifrevisionmarked
\newcommand{\dl}[1]{#1}
\newcommand{\rev}[1]{#1}
\newcommand{\reviewrev}[1]{#1}
\newcommand{\comparisonrev}[1]{#1}
\DeclareRobustCommand{\supprev}[1]{\textcolor{black}{\hypersetup{linkcolor=black}#1}}

\newcommand{\papertitle}{MISeg-Agent: A Human-in-the-Loop Agentic Framework for Error Diagnosis and Executable Prompt Review in 3D Medical Image Segmentation}
\title{\papertitle}
\author{Delin An\textsuperscript{1}, Mingqi Xu\textsuperscript{1},
  Jian-Xun Wang\textsuperscript{1}, and Pengfei Gu\textsuperscript{2}\\[0.2em]
  \normalsize\textsuperscript{1}Cornell University\qquad
  \textsuperscript{2}The University of Texas Rio Grande Valley\\[0.2em]
  \small\ttfamily\hypersetup{urlcolor=black}
  \{\href{mailto:da568@cornell.edu}{da568}, \href{mailto:mx263@cornell.edu}{mx263},
  \href{mailto:jw2837@cornell.edu}{jw2837}\}@cornell.edu\quad
  \href{mailto:pengfei.gu01@utrgv.edu}{pengfei.gu01@utrgv.edu}}
\date{}
\hypersetup{pdftitle={\papertitle},pdfauthor={Delin An, Mingqi Xu, Jian-Xun Wang, Pengfei Gu}}

\begin{document}
\maketitle
\input{sec/0_abstract}
\input{sec/1_introduction}

\input{sec/2_method}
\input{sec/3_experiments}
\input{sec/4_conclusion}
\input{sec/5_ai_statement}

\clearpage
\pdfbookmark[1]{References}{bibliography}
{\small\input{main.bbl}}

\clearpage
\pagenumbering{arabic}
\twocolumn[{\centering\Large
  \pdfbookmark[0]{Supplementary Material}{supplementary-material}
  \textbf{\papertitle}\\
  \vspace{0.5em}Supplementary Material\\
  \vspace{1em}}]
\appendix
\setcounter{table}{0}
\setcounter{figure}{0}
\renewcommand{\thetable}{S\arabic{table}}
\renewcommand{\thefigure}{S\arabic{figure}}
\renewcommand{\theHtable}{S\arabic{table}}
\renewcommand{\theHfigure}{S\arabic{figure}}
\numberwithin{equation}{section}
\renewcommand{\theHequation}{S\thesection.\arabic{equation}}
\input{sec/supplementary}

\end{document}

%% file: sec/0_abstract.tex
\begin{abstract}    
Despite the strong performance of modern medical image segmentation models, their predictions can still contain localized errors that compromise segmentation quality. Identifying and correcting these errors is therefore essential for improving the reliability of existing segmentation results. While interactive segmentation models enable targeted corrections, identifying potential errors, determining where to intervene, and evaluating corrective actions remain challenging. Moreover, each correction changes the current mask, requiring subsequent error diagnosis and recommendations to adapt accordingly. To address these challenges, we propose MISeg-Agent, a human-in-the-loop agentic framework that diagnoses segmentation errors and provides evidence-based correction recommendations. MISeg-Agent first selects a suitable pretrained segmentation model based on benchmark performance and computational requirements. Given its predicted mask, the framework identifies potential missing and excess foreground regions and recommends suspicious slices and corrective points for review. The framework then evaluates candidate point prompts using an interactive segmentation model and presents the resulting corrections for human verification before applying them. By operating directly on images and predicted masks, the error diagnosis process remains independent of the underlying segmentation model architecture. Evaluations on CT datasets demonstrate that MISeg-Agent improves initial segmentation results and produces final masks closely aligned with reference annotations. By integrating adaptive error diagnosis with validated correction recommendations, MISeg-Agent provides an interpretable approach to identifying and reducing residual errors in existing medical image segmentations.

\end{abstract}

%% file: sec/1_introduction.tex
\section{Introduction}
\label{sec:intro}

\begin{figure*}[t]
    \centering

    \begin{minipage}[b]{0.56\textwidth}
        \centering

            \includegraphics[width=\linewidth]{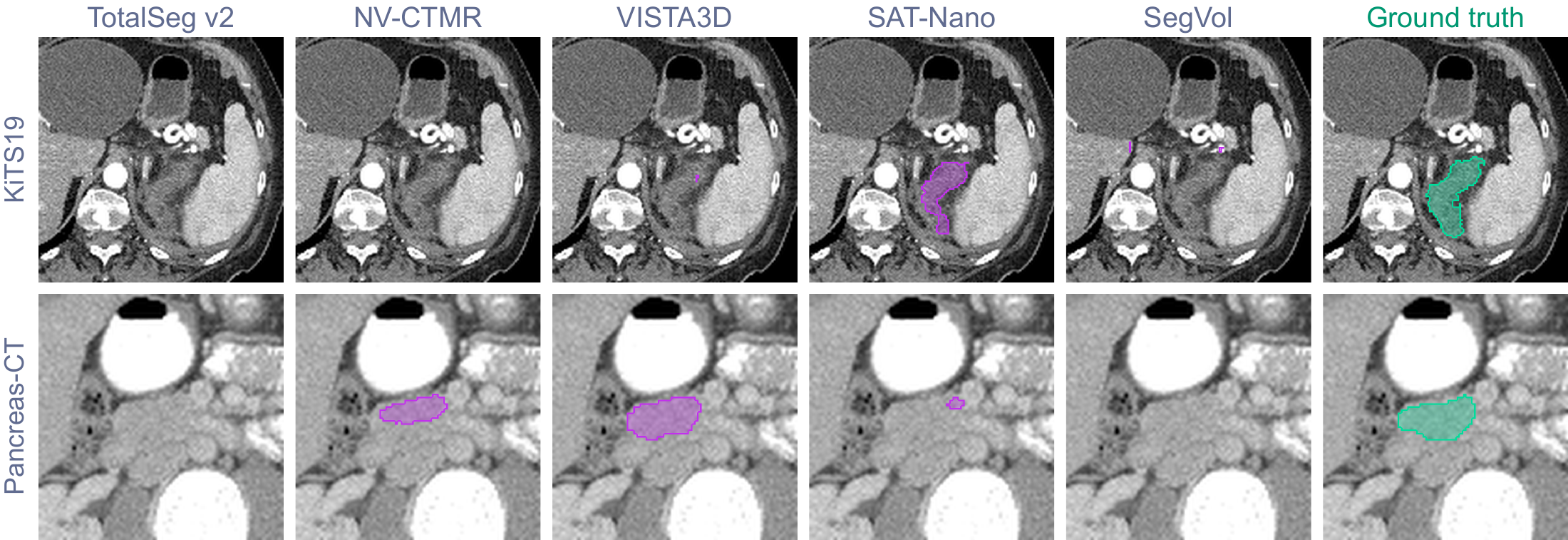}
        \par\smallskip

        (a)
    \end{minipage}
    \hfill
    \begin{minipage}[b]{0.43\textwidth}
        \centering

            \footnotesize
            \renewcommand{\arraystretch}{1.05}
            \setlength{\tabcolsep}{2pt}

            \begin{tabular*}{\linewidth}{@{\extracolsep{\fill}}lccc@{}}
                \toprule
                 & Dice $\uparrow$ & $\Delta$ Dice & Points \\
                \midrule
                Initial mask
                & \dicesummary{0.927}{0.040}
                & $+0.000$
                & 0.0 \\

                Mask only
                & \dicesummary{0.879}{0.124}
                & $-0.048$
                & 0.0 \\
                \midrule

                \reviewrev{All points}
                & \dicesummary{0.701}{0.310}
                & $-0.226$
                & 18.6 \\
                \addlinespace[1pt]

                \reviewrev{First $k$ points}
                & \reviewrev{\dicesummary{0.870}{0.150}}
                & \reviewrev{$-0.057$}
                & \reviewrev{1.7} \\
                \addlinespace[1pt]

                \reviewrev{Screened points}
                & \reviewrev{\dicesummary{0.892}{0.094}}
                & \reviewrev{$-0.035$}
                & \reviewrev{1.7} \\
                \bottomrule
            \end{tabular*}
        \par\smallskip

        (b)
    \end{minipage}

    \caption{
    Motivation for MISeg-Agent.
    (a) Segmentation performance across pretrained models and CT targets, illustrating the variability of initial segmentation quality.
    (b) Mean Dice ($\pm$ standard deviation) over 120 cases for the initial mask and four nnInteractive~\cite{isensee2025nninteractive} refinement settings. The first-$k$ and screened rows use the same initial mask and per-case point count; both retain the initial mask when that count is zero. $\Delta$ Dice is the displayed mean minus the initial-mask mean. Points is the mean submitted count per volume. Screening details appear in Supplementary.
    These results highlight the importance of evaluating corrective prompts before applying them to an existing segmentation.
    }
    \label{fig:motivation}
\end{figure*}

Segmentation masks support quantitative analysis and annotation in 3D medical imaging~\cite{isensee2021nnu,shaker2022unetr++,zhou2021nnformer,hatamizadeh2022swin,hatamizadeh2022unetr,milletari2016v}.
Recent pretrained segmentation models have achieved strong performance and can provide high-quality initial predictions~\cite{he2024vista3d,zhao2023sat,du2023segvol,wasserthal2023totalsegmentator}.
Nevertheless, strong overall accuracy does not guarantee error-free segmentations.
Localized errors, such as missing foreground, leakage into neighboring tissues, and fragmented structures, can persist across different slices and may be overlooked by volume-level evaluation metrics.
Meanwhile, segmentation performance varies across anatomical targets, as do runtime and GPU memory requirements (Fig.~\ref{fig:motivation}(a)).
Therefore, beyond selecting a suitable pretrained model, it is essential to identify residual errors in its predictions and provide targeted recommendations for improving segmentation quality.

Promptable and interactive segmentation methods support target delineation~\cite{SAM,MedSAM,zhang2021interactive} and refinement using spatial prompts~\cite{wang2018interactive,wong2024scribbleprompt,isensee2025nninteractive}.
However, these models do not inherently identify errors in existing segmentations or determine which corrective interactions will improve the current mask.
A positive point intended to recover missing tissue may introduce leakage, while a negative point may remove valid foreground along with false positives.
As shown in Fig.~\ref{fig:motivation}(b), applying all generated point candidates reduces mean Dice from $0.927$ to $0.701$, even below mask-only refinement ($0.879$).
At matched point counts, screened candidates outperform unscreened ones ($0.892$ versus $0.870$), although both remain below the initial-mask Dice.
These observations highlight that corrective prompts do not necessarily improve segmentation quality.
Effective refinement therefore requires identifying potential errors, evaluating candidate corrections, and verifying their effects before updating the segmentation.

Uncertainty-aware models estimate prediction confidence~\cite{Fuchs2022}, while quality-control methods localize potential errors from images and segmentation masks~\cite{SpecktorFadida2025,Zaman2023}.
However, identifying a suspicious region does not directly determine how to correct it or whether a proposed correction will improve the segmentation.
Meanwhile, MedSAM-Agent formulates interactive segmentation as autonomous multi-turn decision-making through reinforcement learning~\cite{Liu2026}.
Our work focuses on diagnosing residual errors in existing segmentations and translating these findings into actionable correction recommendations.
By connecting FN/FP diagnosis across heterogeneous mask sources with slice prioritization, corrective point proposal, and model-response verification, MISeg-Agent provides users with diagnostic evidence and tested correction options while preserving human control over mask updates.

Diagnosing and correcting segmentation errors are interdependent. Once a correction is accepted, the updated mask may contain different errors, making previous diagnoses and recommendations outdated. Each correction must therefore be evaluated against the current mask, and accepted updates require renewed diagnosis. Rejected corrections leave the mask unchanged, while modified prompts require retesting. These dependencies motivate an agentic framework that coordinates diagnostic and interactive tools through a shared case state, keeping recommendations consistent with the latest mask and user feedback.

We introduce MISeg-Agent, a human-in-the-loop agentic framework designed to diagnose residual errors in pretrained segmentation predictions and provide targeted correction recommendations.
The framework first selects a suitable pretrained model based on benchmarked segmentation accuracy, runtime, and GPU memory requirements.
Given the image and current segmentation mask, MISeg-Agent identifies potential FN/FP regions and recommends suspicious slices and corrective points for review.
Its Executable Review Agent evaluates candidate point prompts using nnInteractive~\cite{isensee2025nninteractive}, comparing their responses with the current mask and a mask-only output.
Users can then select points to apply jointly and inspect the resulting segmentation before approving an update.
Accepted updates trigger renewed error diagnosis and correction recommendations.

To evaluate error diagnosis across segmentation models, we build a benchmark with reference annotations and predicted masks from multiple pretrained models.
During inference, MISeg-Agent performs diagnosis using only the medical image and current segmentation mask, without requiring access to the source model's identity or internal features.
We use reference annotations for supervision and evaluation rather than as inputs to the diagnostic process.
We therefore describe the diagnosis interface as \emph{model-agnostic}, indicating its independence from the underlying segmentation architecture rather than guaranteed generalization to arbitrary unseen models.

Our contributions are summarized as follows:
(1) We introduce an agentic framework for diagnosing residual errors in existing medical image segmentations and providing targeted correction recommendations. The diagnosis operates directly on images and predicted masks, supporting multiple pretrained segmentation models without accessing their internal features.
(2) We develop a human-in-the-loop correction workflow that integrates error-guided slice and point recommendations with interactive model-response verification. The framework continuously adapts its diagnosis and recommendations to the current segmentation and user-approved updates.
(3) We construct a cross-model error diagnosis benchmark and evaluate MISeg-Agent on 120 cases from seven CT datasets, including 360 participant--case sessions from three participants. We assess error-diagnosis performance, final segmentation quality, and user review activity.

%% file: sec/2_method.tex
\section{Method: MISeg-Agent}
\label{sec:method}

\begin{figure*}[htbp]
    \centering
    \includegraphics[width=0.91\linewidth]{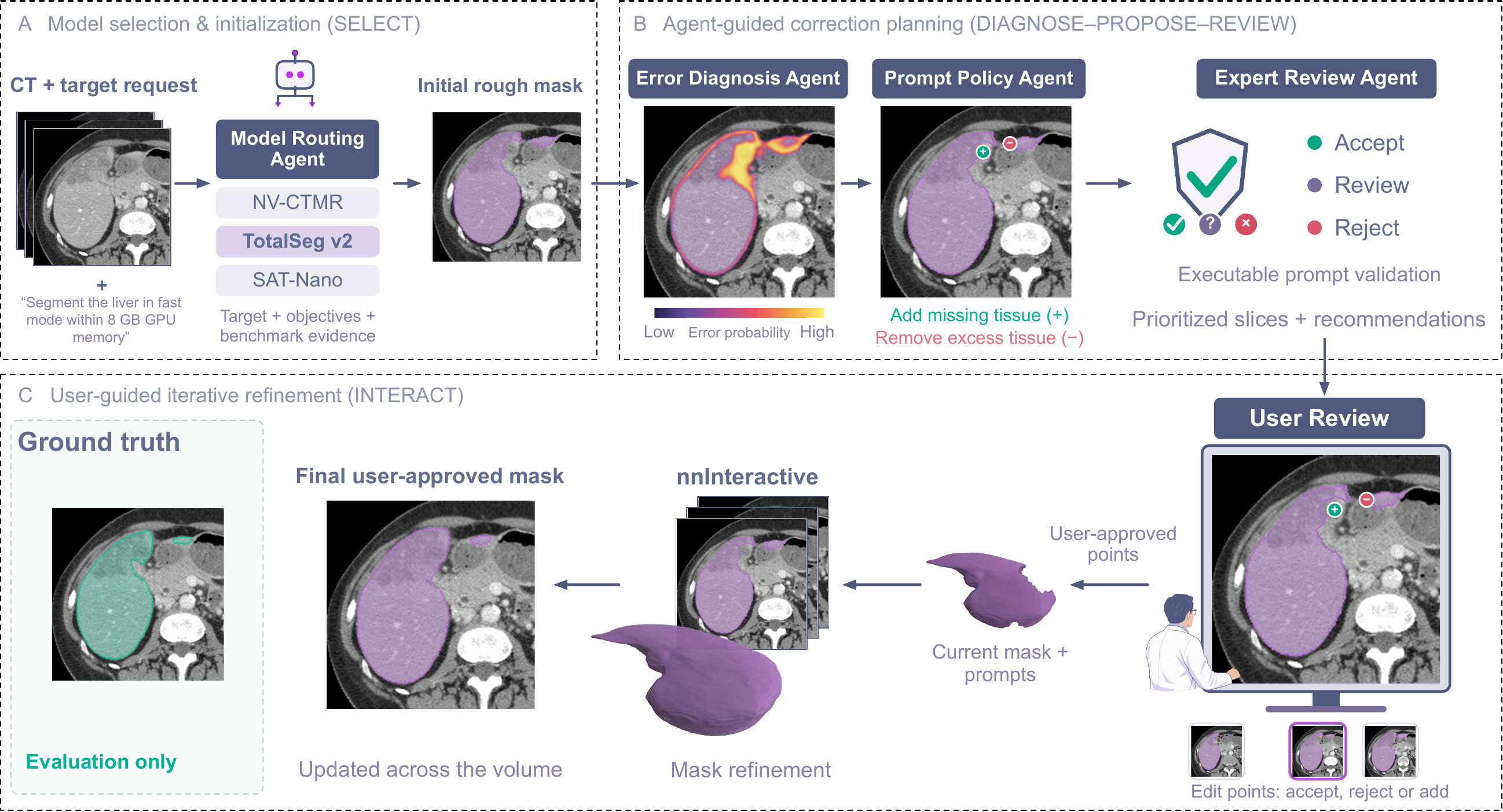}
    \caption{
Overview of MISeg-Agent.
Given a volume and target request, MISeg-Agent selects a compatible \reviewrev{pretrained model} to produce an initial mask, diagnoses FN/FP errors, proposes candidate points, reviews them with nnInteractive, and commits only user-approved refinements. \reviewrev{``Expert'' denotes the Executable Review Agent.}
}
    \label{fig:overview}
\end{figure*}

Let $X\in\mathbb{R}^{N_z\times N_y\times N_x}$ be a volume and $q$ a target request.
MISeg-Agent aims to obtain a final user-approved mask $M^\star$ \dl{through benchmark-informed model selection and targeted error correction.
It first selects a compatible \reviewrev{pretrained segmentation model} to produce an initial rough mask $M^0$, then directs human review to suspected errors and proposed corrections, \reviewrev{providing explicit locations and candidate interactions for user inspection}.}
\reviewrev{Fig.~\ref{fig:overview} summarizes the \textsc{Select--Diagnose--Propose--Review--Interact} workflow.
Selection considers target compatibility, benchmarked segmentation quality, runtime, and GPU memory; diagnosis and proposal identify suspected errors and candidate interventions; executable review assesses their responses before human-controlled refinement.}

\reviewrev{The Interactive Coordinator maintains the current mask version, diagnosis maps, candidate points, probe results, and interaction history.
Here, \emph{agentic} coordination involves invoking tools and selecting candidate points according to the current case state. Uncertain candidates are referred to the user, who controls point submission and mask updates.
The diagnostic predictor provides evidence, proposal and review policies select recommendations, and the user controls point submission and final acceptance.
The MISeg-Agent retests edited points and reruns error diagnosis after each user-approved mask update. \comparisonrev{The agent decides whether to continue refinement or stop; the user retains control over all mask updates.}
Supplementary Section~\ref{sec:agent-contract} summarizes these rules for state updates and user approval.}

\subsection{Benchmark-Informed \reviewrev{Model} Routing}
\label{sec:foundation-selection}
\textcolor{black}{Candidate models.
Let $\mathcal{M}$ index a collection of pretrained segmentation models ${F_m\in\mathcal{M}}$, including both general-purpose and task-specific models.
Our library contains NV-Segment-CTMR, NV-Segment-CT/VISTA3D~\citep{he2024vista3d}, SAT-Nano~\citep{zhao2023sat}, SegVol~\citep{du2023segvol}, and TotalSegmentator~v2~\citep{wasserthal2023totalsegmentator}. 
These models are selected for initialization because they produce target-specific 3D CT masks under fixed inference settings without user-provided spatial prompts. 
We fix their checkpoints, target mappings, and inference settings before evaluation, and filter candidates using fixed compatibility criteria, including availability, dimensionality, CT compatibility, target coverage, and output-mask compatibility.}

\noindent\textbf{Benchmark-informed selection.}
\dl{Compatible models differ in segmentation quality, runtime, and GPU memory usage.}
\reviewrev{Deployment can impose limits on runtime and GPU memory; we therefore include these measured costs with segmentation quality as selection criteria.}
\dl{Routing uses target-specific benchmark evidence to adapt model selection to these competing objectives, allowing the same framework to accommodate different requirements.}
Let $c$ denote the structured target parsed from $q$, $o$ the selection objective, and $\mathcal{M}_c\subseteq\mathcal{M}$ the compatible candidate set.
\reviewrev{The routing context $r$ holds an optional dataset hint and a stable case identifier for deterministic tie-breaking.}
Model selection and execution are expressed as
\reviewrev{$m_0=\mathcal{R}(c,o,\mathcal{M}_c,r;\mathcal{S})$} and
\reviewrev{$M^0=F_{m_0}(X,c)$}, where $\mathcal{R}$ is the routing policy and $\mathcal{S}$ contains the model capabilities and benchmark evidence.

\reviewrev{For compatible models, the benchmark reports mean and 10th-percentile Dice, runtime, and GPU memory use.
A compatible model selected by the user takes precedence; otherwise, accuracy prioritizes quality and then lower-tail Dice, while balanced selection considers lower-tail performance and resource use among models in the highest quality grade.
Fast and low-memory settings minimize the requested resource within that grade unless the user permits an unrestricted choice.
This quality constraint avoids treating a reduction in resource use as sufficient reason to select a lower-grade model by default.
Further routing analysis is provided in Supplementary Section~\ref{sec:routing-details}.}

\subsection{Error Diagnosis Benchmark}
\label{sec:error-benchmark}

\dl{Targeted correction requires identifying likely errors before users search for them.
Because routing can select different \reviewrev{pretrained segmentation models}, diagnosis must operate on heterogeneous masks without access to the source model's internals.
\reviewrev{We therefore evaluate \emph{model-agnostic voxel-wise error diagnosis} as a component of the correction workflow.}}
Given an image volume $X$ and an imperfect binary mask $\widehat{M}$ \dl{produced by a black-box segmentation model}, the goal is to predict for each voxel whether it is correct, FN, or FP with respect to the GT mask $Y$.

We treat each requested structure as a binary segmentation task.
For multi-organ requests, we can apply the procedure independently to each target or to a user-specified foreground label.
For each case, we construct diagnosis targets from the pair $(Y,\widehat{M})$.
The binary error target is
$
E = Y \oplus \widehat{M},
$
where \dl{$\oplus$ denotes the voxel-wise exclusive OR operation,} and $E(v)=1$ indicates that voxel $v$ is incorrectly labeled in $\widehat{M}$.
To distinguish missing foreground from excessive foreground, we define the FN polarity target
\dl{
$
B = Y\odot(1-\widehat{M}),
$}
\dl{with polarity supervision restricted to error voxels where $E=1$.
Here, $\odot$ denotes element-wise multiplication.}
Thus, error voxels with $B=1$ correspond to FNs, while error voxels with $B=0$ correspond to FPs.
\dl{This separation supports two correction decisions: error localization identifies where review is needed, while polarity indicates whether to add or remove foreground.}

\dl{We construct the benchmark by running a fixed library of candidate \reviewrev{pretrained segmentation models} on each dataset--target group, producing the collection}
\begin{equation}
\begin{aligned}
\mathcal{D}_{\mathrm{ED}}
=\bigl\{&(X_i,Y_i,\widehat{M}_{i,m},m,g_i):\; i=1,\ldots,N,\;\\
&m\in\mathcal{M}_{g_i}\bigr\},
\end{aligned}
\end{equation}
where $m$ denotes the model that generated the rough mask, $g_i$ denotes the dataset--target group, and $\mathcal{M}_{g_i}$ is the set of compatible models for that group.
We retain the model identity for splitting and subgroup evaluation but do not provide it as an input to the error diagnosis model.
\dl{Diagnosis therefore relies on the image and current mask, matching the information available after \reviewrev{pretrained model} routing.}

\dl{To assess how the training mask source affects diagnosis, we use rough masks from three representative \reviewrev{pretrained segmentation models}, denoted by $\mathcal{M}_{\mathrm{ED}}$.
For each source model $m\in\mathcal{M}_{\mathrm{ED}}$, we train a separate error diagnosis model using masks generated by that source.
All models are evaluated on the same held-out cases with rough masks from all three sources.
This common evaluation enables comparison of training sources across heterogeneous segmentation outputs.}
\reviewrev{We evaluate binary error localization and FN/FP identification.
These voxel-level metrics assess diagnostic overlap, rather than directly measuring the recall of prioritized slices or the quality of subsequent corrections.}

\subsection{Error Diagnosis Agent}
\label{sec:error-diagnosis}

\reviewrev{The Error Diagnosis Agent estimates voxel-wise correctness and FN/FP scores with a residual 3D U-Net~\cite{Ronneberger2015,Cicek2016}. Its input tensor $Z$ combines two representations of the same CT volume, one emphasizing soft tissue and the other retaining a broader range of intensities, with the current mask and its signed distance map (Supplementary Section~\ref{sec:ed-training}). The scores guide error prioritization and point polarity: positive points target likely missing foreground, and negative points target likely excess foreground. Because the target request is not an input, the model relies on the current mask to indicate which anatomy is being corrected.}

The model maps $Z$ to error, FN-polarity, and auxiliary foreground logits $(z_e,z_b,z_g)$.
\reviewrev{Positive temperature parameters $(T_e,T_b)$ rescale the logits; the probability notation below denotes model estimates, not an empirical calibration guarantee.}
\reviewrev{We factorize the diagnosis scores into overall error score and conditional FN polarity:}
\begin{equation}
\begin{gathered}
e=\sigma(z_e/T_e),\quad \beta=\sigma(z_b/T_b),\\
(\pi_{\mathrm{cor}},\pi_{\mathrm{FN}},\pi_{\mathrm{FP}})
=\bigl(1-e,e\beta,e(1-\beta)\bigr).
\end{gathered}
\label{eq:ed-probabilities}
\end{equation}
\reviewrev{Here, $\sigma$ is the logistic function; $e$ and $\beta$ parameterize normalized three-class scores before candidate filtering.
For binary masks, true error polarity is determined by the current mask once an error is known: FN outside and FP inside.
The learned polarity head supplies a soft score, while proposal support enforces these mask-consistent directions.
The workflow uses these scores without assuming calibration or a benefit from learned polarity.}

\reviewrev{Training supervises three tasks: error localization with binary cross-entropy and focal Tversky losses, FN/FP polarity on error voxels with class-balanced binary cross-entropy, and auxiliary foreground prediction with binary cross-entropy and soft Dice losses. The auxiliary head is used only during training and discarded at inference; Supplementary Section~\ref{sec:ed-training} describes the training objective and loss terms.}

\subsection{Diagnosis-Conditioned Prompt Proposal}
\label{sec:prompt-policy}

\dl{The Prompt Policy Agent converts voxel-wise \reviewrev{diagnostic scores} into explicit choices of review slices, point locations, and prompt labels.
Positive points target predicted FN regions to recover missing foreground, while negative points target predicted FP regions to remove excess foreground.}

\reviewrev{At round $t$, the policy identifies FN candidates outside the current foreground and FP candidates inside it, retaining voxels that satisfy score and margin thresholds:}
\begin{equation}
\begin{aligned}
\Omega_{\mathrm{FN}}^t
&=
\bigl\{v:M^t(v)=0,\;\pi_{\mathrm{FN}}^t(v)\geq\tau,\;\\
&\qquad\pi_{\mathrm{FN}}^t(v)-\pi_{\mathrm{cor}}^t(v)\geq\delta\bigr\},\\
\Omega_{\mathrm{FP}}^t
&=
\bigl\{v:M^t(v)=1,\;\pi_{\mathrm{FP}}^t(v)\geq\tau,\;\\
&\qquad\pi_{\mathrm{FP}}^t(v)-\pi_{\mathrm{cor}}^t(v)\geq\delta\bigr\}.
\end{aligned}
\label{eq:error-support}
\end{equation}
\reviewrev{Here, $v$ indexes voxels, $\tau$ is an error-score threshold, and $\delta$ is the score margin over the correct class.}

\reviewrev{For each 18-connected component $C\subset\Omega_l^t$ containing at least four voxels, where $l\in\{\mathrm{FN},\mathrm{FP}\}$, the evaluated policy selects one interior representative:}
{\ifrevisionmarked\color{black}\fi
\begin{equation}
\begin{aligned}
v_C^*
&=
\arg\max_{v\in C}
\bigl[
\lambda_p\pi_l^t(v)
\\[-1pt]
&\qquad+
\lambda_2\bar d_{2\mathrm{D}}(v,C)
+
\lambda_3\bar d_{3\mathrm{D}}(v,C)
\bigr],
\end{aligned}
\label{eq:prompt-location}
\end{equation}}
\reviewrev{where $d_{2\mathrm{D}}(v,C)$ is the interior distance in the component's axial cross-section through $v$, and $d_{3\mathrm{D}}(v,C)$ is its 3D interior distance.
Both use physical voxel spacing and are normalized separately by their maxima over $C$.
The weights $(\lambda_p,\lambda_2,\lambda_3)=(0.55,0.30,0.15)$ balance diagnostic score and interior depth.
The selected point's axial coordinate determines its review slice.}
\reviewrev{The candidate set $\mathcal C^t$ contains all remaining representatives after suppressing recorded candidate identifiers and points less than 12\,mm from a same-label point in the feedback or refinement history.
Every remaining representative enters executable review; Supplementary Section~\ref{sec:current-policy} summarizes the ordering and history rules.
An empty set returns control to the user for inspection or a recommendation to stop, without certifying an error-free mask.}

\subsection{Executable Prompt \reviewrev{Review}}
\label{sec:prompt-validation}

\dl{Error diagnosis identifies where a mask may be wrong, but does not establish how the interactive model will respond to a proposed correction.
\reviewrev{The Executable Review Agent (abbreviated ``Expert'' in figures and tables) screens candidate points by using nnInteractive to test each candidate point in 3D.
``Expert'' is a software-module label; its verdict is a screening result based on the model response, not an anatomical assessment by a clinician.}}

For candidate point $p_i=(v_i,y_i)$, where $v_i$ is the point location and $y_i\in\{+,-\}$ is the prompt label, we compare its mask-conditioned prediction with a mask-only control:
\begin{equation}
\begin{aligned}
\overline{M}^t&=\mathcal{G}_{\mathrm{nnI}}(X,M^t,\varnothing),\\
\widetilde{M}_i^t&=\mathcal{G}_{\mathrm{nnI}}(X,M^t,\{p_i\}).
\end{aligned}
\label{eq:expert-probe}
\end{equation}
\reviewrev{Each candidate is probed separately from the same current mask.
The two reference masks answer different questions: $M^t$ is the current mask being edited, while $\overline M^t$ shows the refinement model's response without the candidate.
For binary masks viewed as voxel arrays, the signed change decomposes exactly as}
{\ifrevisionmarked\color{black}\fi
\begin{equation}
\widetilde M_i^t-M^t
=\underbrace{\overline M^t-M^t}_{\text{mask-only change}}
+\underbrace{\widetilde M_i^t-\overline M^t}_{\text{control-referenced change}}.
\label{eq:response-decomposition}
\end{equation}}
\reviewrev{This identity explains why both references are retained.
A comparison only with $M^t$ combines mask-only and candidate-conditioned changes; a comparison only with $\overline M^t$ can overlook damage already introduced by mask-only refinement.
We therefore examine both the full output and the control-referenced additions $A_i^t=\widetilde M_i^t\setminus\overline M^t$ and removals $R_i^t=\overline M^t\setminus\widetilde M_i^t$, treating masks as sets in these expressions.
These are observed output differences under the probe configuration, not a decomposition of GT Dice or a causal estimate of prompt benefit.}

\reviewrev{Review uses both this incremental comparison and the full candidate output.
\emph{Local responsiveness} measures agreement with the point label in its neighborhood.
\emph{Directional support} assesses FN evidence on incremental additions for positive points and FP evidence on incremental removals for negative points; positive-point additions are restricted to the candidate-containing component.
\emph{Foreground preservation} limits incremental loss of diagnosis-supported foreground; negative points also require retention relative to the current mask.
A candidate is not accepted solely because it passes the incremental-support checks; it must also pass the applicable full-mask checks.
Structural checks assess the response in 3D; positive points also receive a point-only probe to examine the responding component relative to the current foreground.
Supplementary Sections~\ref{sec:review-evidence} and~\ref{sec:current-policy} summarize the decision rules and main checks.}

\dl{\reviewrev{An automatic \texttt{accept} verdict means that the candidate passes the configured checks for the current image, mask, and diagnosis.}
\reviewrev{These checks reuse the diagnosis that generated the candidate and can inherit its errors; they are not independent evidence of anatomical correctness and do not guarantee the joint effect of accepted points.}
The user therefore previews the joint refinement before approving a mask update.
\reviewrev{The module assigns \texttt{accept}, \texttt{reject}, or \texttt{review}; the last defers a candidate for user inspection.
Unavailable evidence is recorded separately.
These screening verdicts are distinct from user approval of points and commitment of the joint output.}}

\subsection{User Review and Mask Updates}
\label{sec:human-loop}

\reviewrev{MISeg-Agent uses guidance from diagnosis, point proposal, and executable review to focus human inspection while preserving user authority over anatomical decisions.
The interface presents the image, current mask, FN/FP overlays, prioritized slices, candidate points, and probe evidence, helping users navigate to suspected errors and assess proposed actions in their anatomical context.
Users determine whether a highlighted region requires correction and whether the proposed point has an appropriate location and label.
Agent suggestions can be inaccurate or incomplete, so users may accept, reject, relabel, or move candidates and add points for errors not covered by the proposals.
An edited point is treated as an explicit human override and requires a new executable probe before it can be considered reviewed.}

\reviewrev{User-approved points are applied jointly with nnInteractive, and users inspect the resulting mask before accepting or rejecting the update.
This review assesses their combined effect, which may differ from the responses to individual points.
Rejection retains the current mask; acceptance commits the update.
Diagnosis and recommendations are then refreshed for the new mask before further automated guidance is provided.
Supplementary Section~\ref{sec:human-protocol} provides further details of the interaction workflow.}

%% file: sec/3_experiments.tex
\section{Experiments and Results}
\label{sec:exp}

\begin{table*}[t]
\caption{Segmentation Dice across seven CT datasets.
\reviewrev{Cells show mean Dice $\pm$ standard deviation; standard deviations use smaller type.
The original MISeg-Agent result summarizes 360 participant--case sessions; automatic and point-only results summarize 120 cases.
The original workflows differ in initialization and human input; the matched comparison below uses shared initial masks and point budgets.
NV-CTMR: NV-Segment-CTMR; VISTA: NV-Segment-CT/VISTA3D; Panc.: pancreas.}}
\label{tab:segmentation}
\centering
\small
\renewcommand{\arraystretch}{1.03}
\setlength{\tabcolsep}{2pt}
\begin{tabularx}{\linewidth}{@{}>{\raggedright\arraybackslash}p{0.19\linewidth}*{8}{>{\centering\arraybackslash}X}@{}}
\toprule
Model & \shortstack{Aorta\\Seg24} & KiTS19 & \shortstack{MSD\\Liver} & \shortstack{MSD\\Panc.} & \shortstack{MSD\\Spleen} & \shortstack{Panc.\\CT} & \shortstack{SLIVER\\07} & All \\
\midrule
Cases & 13 & 28 & 18 & 39 & 7 & 12 & 3 & 120 \\
\multicolumn{9}{@{}l}{\reviewrev{\emph{Automatic initialization}}} \\
TotalSeg. v2 & \dicesummary{0.871}{0.040} & \dicesummary{0.855}{0.120} & \dicesummary{0.968}{0.011} & \dicesummary{0.722}{0.196} & \dicesummary{0.952}{0.009} & \dicesummary{0.755}{0.174} & \dicesummary{0.968}{0.007} & \dicesummary{0.829}{0.166} \\
NV-CTMR & \dicesummary{0.837}{0.051} & \dicesummary{0.918}{0.094} & \dicesummary{0.970}{0.009} & \dicesummary{0.912}{0.021} & \dicesummary{0.963}{0.007} & \dicesummary{0.877}{0.016} & \dicesummary{\textbf{0.973}}{\textbf{0.007}} & \dicesummary{0.915}{0.063} \\
SAT-Nano & \dicesummary{0.818}{0.040} & \dicesummary{0.952}{0.014} & \dicesummary{0.965}{0.009} & \dicesummary{0.875}{0.032} & \dicesummary{0.957}{0.005} & \dicesummary{0.841}{0.066} & \dicesummary{0.967}{0.003} & \dicesummary{0.904}{0.062} \\
VISTA & \dicesummary{0.859}{0.038} & \dicesummary{0.885}{0.107} & \dicesummary{0.971}{0.009} & \dicesummary{0.871}{0.051} & \dicesummary{0.960}{0.004} & \dicesummary{0.864}{0.022} & \dicesummary{\textbf{0.973}}{\textbf{0.005}} & \dicesummary{0.895}{0.073} \\
SegVol & \dicesummary{0.844}{0.057} & \dicesummary{0.496}{0.399} & \dicesummary{0.946}{0.050} & \dicesummary{0.796}{0.127} & \dicesummary{0.959}{0.007} & \dicesummary{0.814}{0.038} & \dicesummary{\textbf{0.973}}{\textbf{0.005}} & \dicesummary{0.769}{0.262} \\
\midrule
\multicolumn{9}{@{}l}{\reviewrev{\emph{Unfiltered points; no initial mask or human editing}}} \\
ScribblePrompt & \dicesummary{0.004}{0.005} & \dicesummary{0.070}{0.041} & \dicesummary{0.030}{0.028} & \dicesummary{0.014}{0.024} & \dicesummary{0.012}{0.025} & \dicesummary{0.006}{0.006} & \dicesummary{0.033}{0.056} & \dicesummary{0.028}{0.037} \\
nnInteractive & \dicesummary{0.072}{0.161} & \dicesummary{0.608}{0.284} & \dicesummary{0.734}{0.369} & \dicesummary{0.197}{0.187} & \dicesummary{0.306}{0.428} & \dicesummary{0.120}{0.235} & \dicesummary{0.643}{0.554} & \dicesummary{0.369}{0.364} \\
\midrule
\multicolumn{9}{@{}l}{\reviewrev{\emph{Human-guided complete workflow}}} \\
MISeg-Agent & \dicesummary{\textbf{0.916}}{\textbf{0.150}} & \dicesummary{\textbf{0.955}}{\textbf{0.130}} & \dicesummary{\textbf{0.972}}{\textbf{0.008}} & \dicesummary{\textbf{0.931}}{\textbf{0.180}} & \dicesummary{\textbf{0.965}}{\textbf{0.008}} & \dicesummary{\textbf{0.920}}{\textbf{0.190}} & \dicesummary{0.972}{0.008} & \dicesummary{\textbf{0.943}}{\textbf{0.144}} \\
\midrule
\multicolumn{9}{@{}l}{\comparisonrev{\emph{Matched human correction: same initial masks and point budget}}} \\
\comparisonrev{Human + nnInteractive} & \comparisonrev{\dicesummary{0.885}{0.172}} & \comparisonrev{\dicesummary{0.910}{0.160}} & \comparisonrev{\dicesummary{0.961}{0.031}} & \comparisonrev{\dicesummary{0.864}{0.196}} & \comparisonrev{\dicesummary{0.957}{0.018}} & \comparisonrev{\dicesummary{0.848}{0.202}} & \comparisonrev{\dicesummary{0.969}{0.011}} & \comparisonrev{\dicesummary{0.898}{0.163}} \\
\comparisonrev{MISeg-Agent (matched)} & \comparisonrev{\dicesummary{0.912}{0.153}} & \comparisonrev{\dicesummary{0.951}{0.136}} & \comparisonrev{\dicesummary{0.971}{0.010}} & \comparisonrev{\dicesummary{0.924}{0.185}} & \comparisonrev{\dicesummary{0.964}{0.010}} & \comparisonrev{\dicesummary{0.913}{0.193}} & \comparisonrev{\dicesummary{0.972}{0.009}} & \comparisonrev{\dicesummary{0.938}{0.146}} \\
\bottomrule
\end{tabularx}

\end{table*}

\subsection{Datasets and Implementation Details}
We evaluate MISeg-Agent on seven public CT datasets: \dl{AortaSeg24~\cite{AortaSeg24}}, \dl{KiTS19~\cite{Kits19}}, MSD Liver~\cite{simpson2019large}, MSD Pancreas~\cite{simpson2019large}, MSD Spleen~\cite{simpson2019large}, Pancreas-CT~\cite{pancreasct}, and SLIVER07~\cite{van07}, comprising 1,194 cases in total. For training the error-diagnosis model, 1,074 cases are used for training and 120 cases are reserved for \comparisonrev{testing}.
\comparisonrev{These 120 cases are excluded from development-time model selection, hyperparameter tuning, and threshold calibration; the agent selects a model at inference using case characteristics and task requirements, without access to test annotations or cohort-level test performance.}
\reviewrev{We evaluate one binary foreground per volume. KiTS19, MSD Liver, and MSD Pancreas include organ and tumor labels; AortaSeg24 includes main-aorta Zones 0--9 but excludes branch and iliac classes; the other datasets use the organ label.
Supplementary Sections~\ref{sec:target-definitions} and~\ref{sec:ed-training} give the exact label mapping and summarize preprocessing and training settings.
Inference, runtime, and GPU-memory evaluation use the same 120-case \comparisonrev{test} cohort on an NVIDIA GeForce RTX 4090 Laptop GPU with 16 GB of VRAM.
We used \texttt{gpt-5.6-terra} for the language-model component; the framework also accepts other compatible models selected by the user.}

\subsection{Results}
\label{sec:interactive-baselines}
\reviewrev{We evaluate three aspects: final masks under the specified workflows, voxel-wise Dice for predicted error regions, and recorded activity during human interaction.
The 120-case cohort is used for \comparisonrev{testing}; comparisons across the original workflows are descriptive, while the matched comparison controls initialization and point budget.
The Supplementary Material provides the corresponding protocol and evaluation unit for each result.}

\noindent\textbf{Segmentation Performance.}
\dl{\reviewrev{The five pretrained initializers run under fixed inference settings.}
\reviewrev{The ScribblePrompt and nnInteractive baselines receive the original agent-generated points without Expert filtering, human editing, or rough-mask initialization.}
\reviewrev{These point sets contain $20.27\pm30.13$ generated points per case (Supplementary Section~\ref{sec:prompt-statistics}).}
In MISeg-Agent, the framework selects a \reviewrev{pretrained model} to generate the initial mask.
It then refines this mask using reviewed points, with updates committed only after human approval.}

\reviewrev{The strongest initializer in Table~\ref{tab:segmentation} varies across targets.
MISeg-Agent reports $0.943\pm0.144$ overall, compared with $0.915\pm0.063$ for NV-Segment-CTMR, and the highest mean on six of seven datasets.
Differences from the best automatic model are largest on AortaSeg24 and Pancreas-CT and smaller on MSD Liver and Spleen.
SLIVER07 scores are 0.972 for MISeg-Agent and 0.973 for the best baseline.
These describe different workflows, with routing and human-input contributions combined.}

\reviewrev{With unfiltered agent-generated points and no rough-mask prior, ScribblePrompt and nnInteractive obtain $0.028\pm0.037$ and $0.369\pm0.364$, respectively.
Supplementary Section~\ref{sec:participant-point-baselines} provides complementary results with participant-supplied points at specified budgets.}

\noindent\textbf{Matched Human-Guided Comparison.}
To assess the added value of framework assistance, we compared direct human correction using nnInteractive with MISeg-Agent on the same cases, under identical initial masks and matched correction-point budgets.
In the direct condition, users locate errors and place corrective points without error maps, proposed points, or executable screening.
Both conditions use nnInteractive for refinement and require human approval of mask updates.
The budget counts all submitted corrective points, whether placed manually or chosen from recommendations.
The matched rows in Table~\ref{tab:segmentation} report final-mask Dice under these conditions.

\comparisonrev{The matched comparison demonstrates the benefit of the overall assistance workflow. Since nnInteractive generates a complete 3D segmentation, local corrections may inadvertently alter accurately segmented regions. MISeg-Agent mitigates this risk by evaluating candidate responses and presenting their effects for human review before applying mask updates. The larger improvements on anatomically challenging targets highlight the importance of evaluating corrective actions, although this comparison does not isolate verification's individual contribution. In contrast, smaller gains on well-segmented organs indicate limited room for further improvement.
}

\noindent\textbf{\dl{Error Diagnosis Benchmark.}}
\dl{\reviewrev{We compare error-diagnosis models trained on three rough-mask sources under common \comparisonrev{test} inputs.}
Separate Error Diagnosis (ED) models are trained on masks from NV-CTMR, TotalSegmentator, and SegVol, which have descending overall segmentation Dice.
Every ED model is evaluated on the same held-out collection containing masks from all three sources for the same 120 cases.
The model receives only the image and rough mask, without the source model's identity.
\reviewrev{The pooled evaluation combines matched-source and cross-source inputs; each training source remains represented.}}

\begin{table}[t]
\caption{Model-agnostic error-diagnosis performance under different rough-mask training sources (mean $\pm$ standard deviation). TotalSeg.: TotalSegmentator.}
\label{tab:ed}

\centering
\footnotesize
\renewcommand{\arraystretch}{1.03}
\setlength{\tabcolsep}{1.3pt}
\begin{tabularx}{\linewidth}{@{}>{\raggedright\arraybackslash}p{0.20\linewidth}*{4}{C}@{}}
\toprule
\shortstack[l]{ED training\\source} & \shortstack{FN\\Dice} & \shortstack{FP\\Dice} & \shortstack{Macro\\Dice} & \shortstack{Binary\\Dice} \\
\midrule

TotalSeg.
& \dicesummary{0.420}{0.160}
& \dicesummary{0.680}{0.140}
& \dicesummary{0.550}{0.120}
& \dicesummary{0.560}{0.150} \\

NV-CTMR
& \dicesummary{0.440}{0.150}
& \dicesummary{0.680}{0.130}
& \dicesummary{0.560}{0.110}
& \dicesummary{0.570}{0.140} \\

SegVol
& \dicesummary{\textbf{0.550}}{\textbf{0.130}}
& \dicesummary{\textbf{0.750}}{\textbf{0.110}}
& \dicesummary{\textbf{0.650}}{\textbf{0.100}}
& \dicesummary{\textbf{0.660}}{\textbf{0.120}} \\

\bottomrule
\end{tabularx}

\end{table}

\dl{Table~\ref{tab:ed} reports FN Dice and FP Dice for missing and excess foreground, respectively.
Macro Dice averages the two,
$\mathrm{Macro\ Dice} = 0.5(\mathrm{FN\ Dice} + \mathrm{FP\ Dice})$,
while Binary Dice measures their union regardless of polarity.
\reviewrev{They quantify voxel-level error overlap, rather than downstream correction benefit.}}
\reviewrev{Initialization and diagnostic supervision are different selection problems.
The SegVol-trained diagnosis model has the highest values on all four metrics, although SegVol has the lowest segmentation Dice among these sources.
Thus, the observed initialization ranking does not transfer to choosing the training-mask source for error diagnosis in this cohort.
MISeg-Agent treats these as separate choices: routing uses segmentation quality and runtime and GPU-memory measurements, while ED selection uses error-localization performance.
Error prevalence, extent, and target composition remain possible explanations for the observed ranking.}

\noindent\textbf{\reviewrev{Human Interaction and Review Activity.}}
\reviewrev{Three participants independently corrected the same 120 cases without access to reference annotations during interaction: a postdoctoral researcher (A), a doctoral student (B), and an undergraduate researcher with limited familiarity with medical images (C).
All sessions followed the review workflow described in Supplementary Section~\ref{sec:current-policy}.
Table~\ref{tab:users} reports the resulting final-mask accuracy and recorded interaction activity under this assisted annotation setting.}
\rev{The 120 cases contain 30,720 axial slices in total, based on the $256\times256\times256$ volumes used in the interaction interface.
Across this shared case set, Participants A, B, and C explicitly confirmed 741, 654, and 734 reviewed slices, respectively (Supplementary Section~\ref{sec:interaction-accounting}).}
\reviewrev{Across the 360 participant--case sessions, these 2,129 slice confirmations correspond to 5.91 per session and 10.09 per recorded review minute.}

\begin{table*}[t]
\caption{\reviewrev{Participant-level final-mask Dice and recorded interaction activity.
Dice is mean $\pm$ standard deviation over 120 cases per participant or 360 participant--case sessions pooled.
Slices/total reports explicitly confirmed review slices / working-grid slices (30,720 per participant; 92,160 pooled).
Review min records intervals for inspection, point editing, and decisions; Slices/min divides confirmed slices by Review min.
Points are manual additions; Accept and Reject count user decision actions. The Pooled row combines A--C.}}
\label{tab:users}

\centering
\small
\renewcommand{\arraystretch}{1.03}
\setlength{\tabcolsep}{4pt}
\begin{tabular*}{\linewidth}{@{\extracolsep{\fill}}lrrrrrrrr@{}}
\toprule
User & \reviewrev{Sessions} & Dice $\uparrow$ & Points & Accept & Reject & \reviewrev{\shortstack{Slices /\\total}} & \reviewrev{\shortstack{Review\\min}} & \reviewrev{\shortstack{Slices/\\min}} \\
\midrule

\reviewrev{A}
& 120
& \dicesummary{0.957}{0.079}
& 39 & 126 & 327 & \rev{741\,/\,30,720} & 73.54 & 10.08 \\

\reviewrev{B}
& 120
& \dicesummary{0.947}{0.148}
& 24 & 103 & 427 & \rev{654\,/\,30,720} & 61.64 & 10.61 \\

\reviewrev{C}
& 120
& \dicesummary{0.925}{0.185}
& 26 & 232 & 191 & \rev{734\,/\,30,720} & 75.92 & 9.67 \\

Pooled
& 360
& \dicesummary{0.943}{0.144}
& \reviewrev{89} & 461 & 945 & \rev{2,129\,/\,92,160} & 211.10 & 10.09 \\

\bottomrule
\end{tabular*}

\end{table*}

\begin{figure*}[t]
    \centering
    \includegraphics[width=0.91\linewidth]{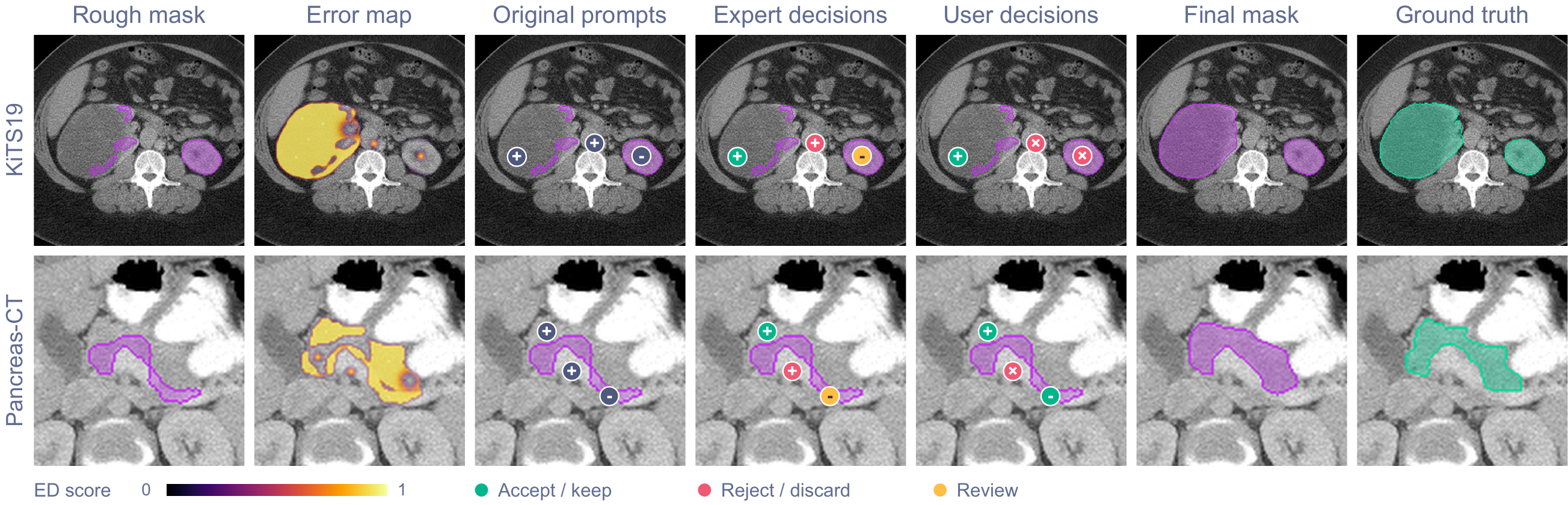}
    \caption{
Qualitative case studies of MISeg-Agent on KiTS19 and Pancreas-CT. Each row shows a selected axial slice. Plus and minus signs denote foreground and background prompts. Green, red, and orange markers indicate accept, reject, and review, respectively; the color bar shows the error-diagnosis (ED) score.
}
    \label{fig:case-study}
\end{figure*}

\rev{%
\noindent\textbf{Annotation-time Context.}
For context, we report annotation times reported in related medical image annotation studies.
For MSD Spleen, MONAI Label~\cite{DiazPinto2022} reports about 25~minutes per volume with a paintbrush tool and 12.5~minutes with a contouring tool.
The IMVS user study~\cite{Reen2026} reports manual annotation times of 8.11 and 73.53~minutes for MSD Pancreas and LiTS, reduced to 5.26 and 47.56~minutes with a copy-and-edit workflow.
Across our 360 participant-case sessions, mean recorded human review and editing time was 35.18 seconds, with dataset-specific means of 19.34, 24.08, and 37.77 seconds for MSD Spleen, MSD Pancreas, and MSD Liver, respectively.
\comparisonrev{Our evaluation focuses on providing error information to help users improve existing masks, so we report interaction time as descriptive context rather than conduct a controlled efficiency comparison.}
\comparisonrev{The recorded times cover review and editing activity, not end-to-end annotation; annotation targets, completion criteria, and timing protocols also differ across studies.}}

\noindent\textbf{Case Study.}
\reviewrev{Fig.~\ref{fig:case-study} illustrates targeted correction in KiTS19 and Pancreas-CT.
In KiTS19, diagnosis highlights a large missed target region.
The panel follows proposed candidates through automated review to the point retained by the user; refinement recovers the missing structure and improves visual agreement with GT.}
\reviewrev{In Pancreas-CT, diagnosis identifies suspected missing and excess foreground for positive and negative point proposals.
Executable review informs user selection, and the approved refinement follows the GT more closely.
These selected examples illustrate the workflow, rather than isolating component contributions or establishing typical-case performance.}

%% file: sec/4_conclusion.tex
\section{Conclusions and Discussion}
\label{sec:conclu}

\reviewrev{MISeg-Agent coordinates 3D mask correction through benchmark-informed model selection, error diagnosis, point proposal, executable review, and user-controlled refinement. It identifies suspected errors in the current mask, tests proposed points with an interactive model, and presents the observed responses for user assessment. \comparisonrev{Further refinement requires renewed diagnosis of the updated mask and an agent decision on whether to continue.} \rev{In the evaluated CT cases, the workflow directed users to localized segmentation errors and supported targeted corrections, resulting in final masks with high overlap with the reference annotations.}}

\comparisonrev{We observed that as mask quality improves, residual error regions become smaller and the error-diagnosis model performs less reliably.
The agent currently judges whether to continue refinement or stop, while mask updates remain subject to user approval.
Improving diagnosis of small residual errors and establishing reliable stopping criteria for iterative refinement are priorities for future work.}

\reviewrev{\rev{The CT findings require validation on independent cohorts and other modalities.} \comparisonrev{The matched comparison assesses assistance as a whole; component-level ablations remain necessary to separate its contributions,} and the three-participant study is too limited to draw strong conclusions about workload or user-dependent effects. Screening may inherit diagnostic errors, while favorable single-point probes do not guarantee the quality of their joint refinement; user preview therefore remains essential. Future work will pursue \comparisonrev{systematic component ablations}, larger controlled user studies, and additional interaction types beyond point prompts.}

%% file: sec/5_ai_statement.tex
\section*{LLM Usage Statement}
The authors used AI-based writing assistance tools only to check grammar, improve clarity, and refine the presentation of the manuscript. 
All technical ideas, method design, experiments, results, analyses, claims, and final writing decisions were reviewed, verified, and approved by the authors. 
AI tools were not used to generate experimental results, fabricate data, or make autonomous scientific conclusions.

%% file: sec/supplementary.tex
\section{Full Routing Analysis}
\label{sec:routing-details}
\rev{The routing policy in main-paper Section~\ref{sec:foundation-selection}
uses benchmark evidence summarized for each dataset--target group $g$.
For a compatible model $m$, quality and \reviewrev{lower-tail Dice} grades are defined by}
\begin{equation}
G^{\mathrm{qual}}_{m,g}=Q(\overline{D}_{m,g}),
\qquad
G^{\mathrm{rel}}_{m,g}=Q(D^{(0.1)}_{m,g}),
\label{eq:tier}
\end{equation}
\rev{where $\overline{D}_{m,g}$ and $D^{(0.1)}_{m,g}$ are the mean and
10th-percentile Dice, respectively.
\reviewrev{For either statistic, $Q(d)$ is weak for $d<0.75$, fair for $0.75\leq d<0.85$, good for $0.85\leq d<0.90$, strong for $0.90\leq d<0.95$, and excellent for $d\geq0.95$.
These grades summarize average and lower-tail overlap, not reliability on a new case.
Dataset-specific grades require a matching target and at least five benchmark cases; otherwise pooled target grades are used.}}

\reviewrev{A compatible user-selected model takes precedence.
Otherwise, accuracy prioritizes quality and then lower-tail grade; balanced keeps the highest quality grade and removes candidates dominated in lower-tail grade, runtime, and memory.
Fast and low-memory objectives minimize the requested resource within the highest quality grade, unless the user permits an unrestricted choice.
Ties are resolved deterministically using model identifiers and the case identifier.
Table~\ref{tab:routing} reports the resulting assignments under the supplied dataset hints.
Routing uses these benchmark grades at inference without access to test annotations.}

\begin{table*}[t]
\caption{\supprev{Model assignments by routing objective. Counts are shown when cases are split between models. Fast and low-memory routing use the highest quality grade. Without that constraint, both select SegVol for all 120 cases. NV-CTMR, VISTA, and TotalSeg.~v2 denote NV-Segment-CTMR, NV-Segment-CT/VISTA3D, and TotalSegmentator v2, respectively.}}
\label{tab:routing}

\centering
\small
\setlength{\tabcolsep}{3pt}
\renewcommand{\arraystretch}{1.03}
\renewcommand{\tabularxcolumn}[1]{m{#1}}
\begin{tabularx}{\linewidth}{@{}lrLLcc@{}}
\toprule
\supprev{Dataset} & \supprev{Cases} & \supprev{Accuracy} & \supprev{Balanced} & \supprev{Fast} & \supprev{Low memory} \\
\midrule
AortaSeg24 & 13 & VISTA~(8), \supprev{TotalSeg.~v2~(5)} & VISTA~(9), \supprev{TotalSeg.~v2~(4)} & \supprev{VISTA} & \supprev{TotalSeg.~v2} \\
KiTS19 & 28 & \supprev{SAT-Nano} & \supprev{SAT-Nano} & \supprev{SAT-Nano} & \supprev{SAT-Nano} \\
MSD Liver & 18 & VISTA~(3), \supprev{NV-CTMR~(3),\newline SAT-Nano~(8), TotalSeg.~v2~(4)} &
\supprev{NV-CTMR~(6), SAT-Nano~(3),\newline SegVol~(4), TotalSeg.~v2~(5)} & \supprev{SegVol} & \supprev{SegVol} \\
MSD Pancreas & 39 & \supprev{NV-CTMR} & \supprev{NV-CTMR} & \supprev{NV-CTMR} & \supprev{NV-CTMR} \\
MSD Spleen & 7 & VISTA~(3), \supprev{NV-CTMR~(4)} & \supprev{NV-CTMR~(2), SegVol~(5)} & \supprev{SegVol} & \supprev{SegVol} \\
Pancreas-CT & 12 & \supprev{NV-CTMR} & \supprev{NV-CTMR} & \supprev{NV-CTMR} & \supprev{NV-CTMR} \\
SLIVER07 & 3 & \supprev{NV-CTMR~(1), SAT-Nano~(2)} & \supprev{NV-CTMR~(1), SegVol~(1), TotalSeg.~v2~(1)} & \supprev{SegVol} & \supprev{SegVol} \\
\bottomrule
\end{tabularx}
\end{table*}

\begin{figure*}[t]
\centering
\includegraphics[width=\linewidth]{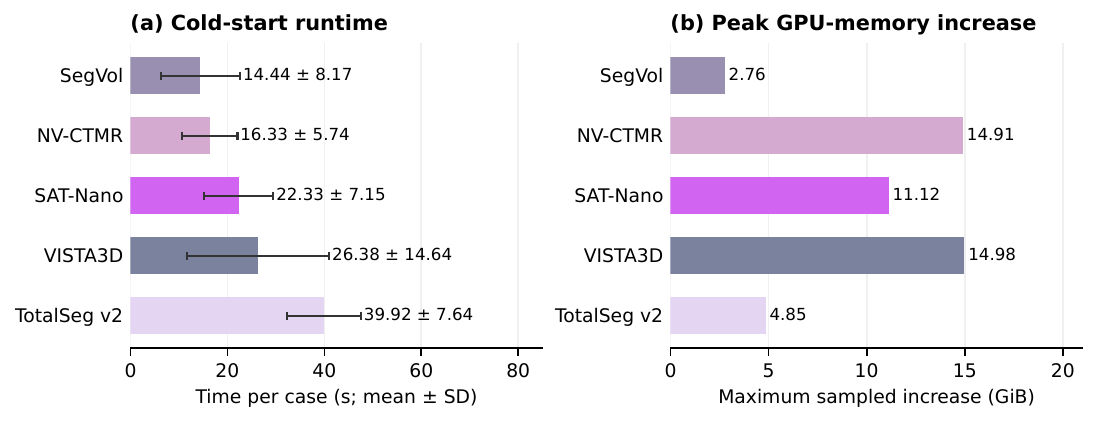}
\caption{Runtime and GPU-memory measurements underlying model routing. Five models evaluated on 120 cases using an RTX 4090 Laptop GPU (16,376 MiB). Left: cold-start wall time, mean $\pm$ std, including loading and the full processing pipeline. Right: maximum sampled GPU-memory increase above baseline across cases under Windows WDDM, not per-process allocation. Lower is better.}\label{fig:routingresources}
\end{figure*}

\reviewrev{Assignments change with both the target and the selection objective; no one model is chosen for every setting.
Fig.~\ref{fig:routingresources} shows the runtime and memory measurements used alongside the quality grades.
When benchmark evidence is insufficient, routing falls back to capability constraints or user selection.}

\section{\reviewrev{Error-Diagnosis Inputs and Training Objective}}
\label{sec:ed-training}
\dl{A 3D residual U-Net combines CT appearance with the current segmentation and its signed distance map as a four-channel input:}
\begin{equation}
\begin{aligned}
Z(X,M)=\bigl[&W_{\mathrm{soft}}(X),\;W_{\mathrm{wide}}(X),\\
&M,\;\phi(M)\bigr],
\end{aligned}
\label{eq:ed-input}
\end{equation}
where $W_{\mathrm{soft}}$ and $W_{\mathrm{wide}}$ are fixed soft-tissue and wide-field CT windowing operators normalized to $[0,1]$.
\dl{The signed distance $\phi\in[-1,1]$ is computed in physical units and clipped at a fixed radius, with negative values inside $M$ and positive values outside.}
\reviewrev{The target request is not part of this input, so targeted diagnosis relies on the current mask indicating the intended anatomy.}

\rev{The targets $E$, $B$, and $Y$ denote binary segmentation error, FN polarity, and foreground ground truth, respectively; $(z_e,z_b,z_g)$ are the corresponding logits.}
\dl{The training objective combines error localization, polarity classification, and auxiliary foreground supervision:}
\begin{equation}
\begin{split}
\mathcal{L}_{\mathrm{total}}={}&
\lambda_e\!\left[
\mathcal{L}_{\mathrm{BCE}}(z_e,E)
+
\mathcal{L}_{\mathrm{FT}}(z_e,E)
\right]\\
&+
\lambda_b
\mathcal{L}_{\mathrm{bal}\text{-}\mathrm{BCE}}(z_b,B;E)\\
&+
\lambda_g\!\left[
\mathcal{L}_{\mathrm{BCE}}(z_g,Y)
+
\mathcal{L}_{\mathrm{Dice}}(z_g,Y)
\right]
\end{split}
\label{eq:ed-loss}
\end{equation}
\dl{where $\mathcal{L}_{\mathrm{BCE}}$ denotes binary cross-entropy loss and $\mathcal{L}_{\mathrm{FT}}$ denotes focal Tversky loss.
The nonnegative weights $(\lambda_e,\lambda_b,\lambda_g)$ sum to one and balance the three supervision objectives.
The auxiliary term includes $\mathcal{L}_{\mathrm{Dice}}$, the soft Dice loss between the foreground prediction $\sigma(z_g)$ and the GT mask $Y$.}
The term $\mathcal{L}_{\mathrm{bal}\text{-}\mathrm{BCE}}$ is class-balanced BCE restricted to voxels where $E=1$.
\dl{The auxiliary foreground head provides anatomical supervision during training and is discarded at inference.
\reviewrev{Training masks are outputs from candidate pretrained models. They represent initialization errors; errors after human edits may follow a different distribution.}}
\reviewrev{The models were trained on volumes resampled to $256\times256\times256$ voxels for 300 epochs using AdamW and cosine learning-rate decay.
Main-paper Table~\ref{tab:ed} evaluates each trained model on masks from all three rough-mask sources, allowing initialization quality and diagnosis-training utility to be compared separately.}

\section{\reviewrev{Review Evidence and Evaluation Scope}}
\subsection{\reviewrev{Evidence validity and user control}}
\label{sec:agent-contract}
\label{sec:human-protocol}
\reviewrev{A mask update changes the input on which later diagnosis and recommendations depend.
The coordinator therefore links each diagnosis to a mask version and each point probe to both that version and the point's location and label.
Editing a point requires a new probe; committing a mask preview requires new diagnosis; rejecting a preview retains the current mask.
If no proposal is eligible, the workflow returns control to the user rather than declaring the mask correct.
}

\reviewrev{Users select the points to submit, including any edited or manually added points.
The interactive model applies the selected points jointly; its combined output is shown before the user commits or rejects the mask update.
Individual point probes do not establish the quality of that joint output.
The interaction record preserves mask revisions, point changes, and user decisions for review.}

\subsection{\reviewrev{What executable review checks}}
\label{sec:review-evidence}
\reviewrev{\supprev{In the participant workflow, nnInteractive runs from the current mask both with and without the candidate point.}
The current mask anchors checks of the complete output, while the mask-only response shows changes produced by the interactive model without the candidate.
Changes between the candidate and mask-only outputs are checked for the requested direction: additions should agree with predicted missing foreground for a positive point, and removals with predicted excess foreground for a negative point.
Local response, preservation of diagnosis-supported foreground, and large structural changes are also checked.
These are observed model responses, not ground-truth accuracy or a causal estimate of a point's benefit.}

The automatic screen can accept a candidate, reject a decisive inconsistency, or defer uncertain evidence for user review; it does not commit a mask.
Screening uses the image, current mask, diagnosis, and model responses, without GT.

\subsection{\reviewrev{Targets and evaluation units}}
\label{sec:evaluation-scope}
\label{sec:target-definitions}
\begin{table*}[t]
\caption{\supprev{Inputs and evaluation units by protocol. The protocols reuse the same 120 test cases. Each session pairs one participant with one case. In the original assisted workflow, three participants each reviewed all 120 cases.}}
\label{tab:protocol-map}
\centering
\small
\setlength{\tabcolsep}{3pt}
\renewcommand{\arraystretch}{1.03}
\begin{tabularx}{\linewidth}{@{}llLl@{}}
\toprule
\supprev{Source} & \supprev{Protocol} & \supprev{Input and intervention} & \supprev{Evaluation unit} \\
\midrule
\supprev{Main Fig.~\ref{fig:motivation}(b)} & \supprev{Automatic refinement} & \supprev{Initial mask and four refinement settings (mask only, all points, first $k$ points, $k$ screened points).} & 120 cases \\
\supprev{Main Table~\ref{tab:segmentation}} & \supprev{Automatic initialization} & \supprev{Fixed pretrained model without points or human edits.} & 120 cases \\
\supprev{Main Table~\ref{tab:segmentation}} & \supprev{Point-only models} & \supprev{Unfiltered agent-generated points without an initial mask or human edits.} & 120 cases \\
\supprev{Main Tables~\ref{tab:segmentation} and~\ref{tab:users}} & \supprev{Original MISeg-Agent} & \supprev{Routed mask, user-selected or added points, and user-approved updates.} & \supprev{360 sessions} \\
\supprev{Main Table~\ref{tab:segmentation}} & \supprev{Matched correction} & \supprev{Direct correction and MISeg-Agent use the same initial masks, point budgets, and nnInteractive refiner.} & \supprev{Sessions} \\
\supprev{Main Table~\ref{tab:ed}} & \supprev{Error diagnosis} & \supprev{Each model trains on one mask source and is tested on all three.} & 120 cases \\
\supprev{Table~\ref{tab:interactive-baselines}} & \supprev{Point-budget baselines} & \supprev{Participant-supplied points without an initial mask.} & 120 cases \\
\bottomrule
\end{tabularx}
\end{table*}

\reviewrev{Each evaluated volume has one binary foreground.
KiTS19, MSD Liver, and MSD Pancreas combine organ and tumor labels (IDs 1 and 2); AortaSeg24 combines main-aorta Zones 0--9 while excluding branch and iliac classes.
MSD Spleen, Pancreas-CT, and SLIVER07 use the organ label.
These definitions apply to overlap evaluation and participant review.}

\reviewrev{Table~\ref{tab:protocol-map} separates results that use different inputs and evaluation units.
Main-paper Fig.~\ref{fig:motivation}(b) contrasts the initial mask with nnInteractive refinement without points, with all generated points, with a count-matched unscreened subset, and with automatically screened points.
\comparisonrev{The original MISeg-Agent results} in main-paper Tables~\ref{tab:segmentation} and~\ref{tab:users} instead summarize the human-guided workflow.
\comparisonrev{The additional matched block in \supprev{main-paper} Table~\ref{tab:segmentation} compares direct human correction with framework assistance under shared initialization and point budgets.}}

\subsection{\reviewrev{Participant activity and timing}}
\label{sec:interaction-accounting}
\label{sec:human-timing}
\reviewrev{\supprev{Main-paper} Table~\ref{tab:users} counts distinct axial slices reviewed within each participant--case session.
Each participant's 120 resampled working-grid volumes contain $120\times256=30,720$ axial slices, and the pooled denominator of $92,160$ repeats the same cases across three participants.
Points are manual additions; Accept and Reject count user decisions, which may cover multiple candidates.
The pooled Dice summaries use 360 participant--case observations on 120 cases.}

\reviewrev{Review min sums recorded intervals for inspection, point editing, and decisions.
The participant rows total 211.10 minutes, or 35.18 seconds per session across 360 sessions; 2,129 slice confirmations divided by 211.10 minutes gives the pooled 10.09 Slices/min in main-paper Table~\ref{tab:users}.
These intervals characterize logged interaction rather than end-to-end completion time.}

\section{\reviewrev{Point-Based Segmentation Baselines}}
\label{sec:prompt-statistics}
\reviewrev{For the point-only baselines in main-paper Table~\ref{tab:segmentation}, the original candidate sets contain 2,432 generated points across 120 cases (806 positive and 1,626 negative; $20.27\pm30.13$ points per case), without rough-mask initialization or human editing.
The complementary experiments below use participant-supplied points.}

\subsection{\reviewrev{Participant-supplied point baselines}}
\label{sec:participant-point-baselines}
\reviewrev{To contextualize the unfiltered-point results, we evaluate ScribblePrompt and nnInteractive on the same 120 cases using points manually placed by participants based on their assessment of each case.
Both models receive CT and point coordinates and labels, without an initial mask.
Table~\ref{tab:interactive-baselines} reports results with one, five, or ten points per volume, and one point per target slice for ScribblePrompt.}

\begin{table}[htbp]
\caption{\supprev{Dice~$\uparrow$ (mean $\pm$ SD) on 120 cases with participant-supplied points. Budgets are fixed per volume or set to one point per target slice ($95.850\pm31.620$ per volume). A dash marks an unevaluated setting.}}
\label{tab:interactive-baselines}
\centering
\small
\renewcommand{\arraystretch}{1.03}
\setlength{\tabcolsep}{2pt}
\begin{tabular*}{\linewidth}{@{\extracolsep{\fill}}lcc@{}}
\toprule
\supprev{Point budget} & ScribblePrompt & nnInteractive \\
\midrule
1 \supprev{per volume} & \dicesummary{0.025}{0.029} & \dicesummary{0.689}{0.233} \\
5 \supprev{per volume} & \dicesummary{0.094}{0.052} & \dicesummary{0.912}{0.057} \\
10 \supprev{per volume} & \dicesummary{0.161}{0.069} & \dicesummary{0.928}{0.041} \\
\midrule
1 \supprev{per target slice} & \dicesummary{0.717}{0.125} & --- \\
\bottomrule
\end{tabular*}
\end{table}

\reviewrev{With participant-supplied points, nnInteractive reaches $0.912$ Dice at five points and $0.928$ at ten, compared with $0.689$ at one.
ScribblePrompt reaches $0.161$ with ten points and $0.717$ under the per-slice protocol, which uses substantially more points.
Thus, the unfiltered-point scores do not characterize model performance under other prompt protocols.
These references differ from MISeg-Agent in initialization and human interaction and do not isolate the benefit of the complete workflow.}

\begin{figure*}[t]
\centering
\includegraphics[width=\linewidth]{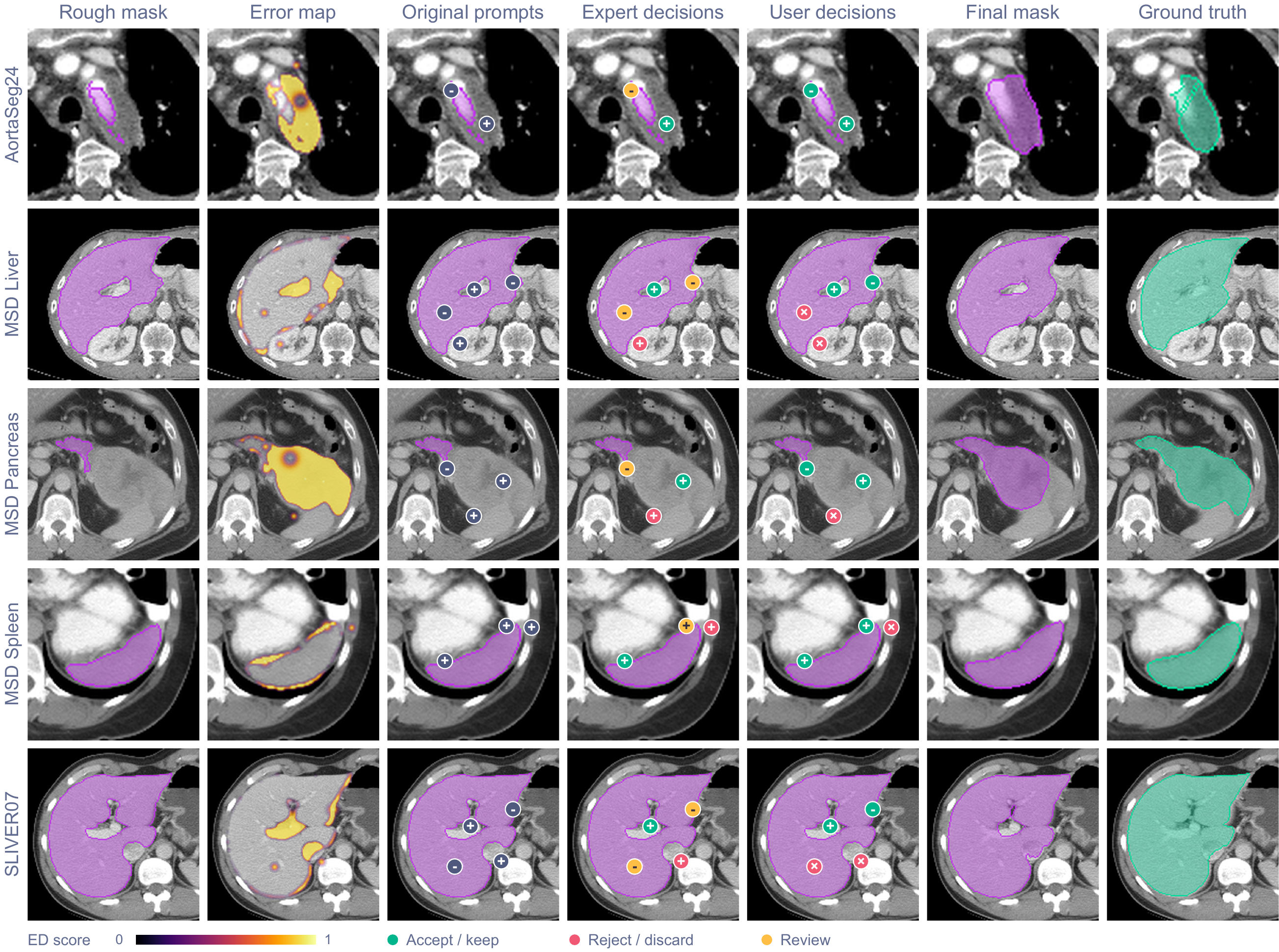}
\caption{\rev{Additional qualitative case studies from the five CT datasets not shown in the main-paper case-study panel.}}
\label{fig:additional-cases}
\end{figure*}

\section{\reviewrev{Screening Context for the Reported Workflows}}
\subsection{\reviewrev{Automatic screen in main-paper Fig.~\ref{fig:motivation}(b)}}
\label{sec:archived-policy}
\reviewrev{The screened row uses a fixed automatic rule applied to candidate points from the same pool as both unscreened rows.
Each point is probed separately from the common rough mask; the rule checks point-label response, whole-mask plausibility, and FN/FP support before accepting it for submission.
Accepted points are then submitted together in their recorded order; if none are accepted, the initial mask is retained.
The all-points and screened rows submit different numbers of points; the matched-budget unscreened row submits the first $k$ candidates in proposal order, where $k$ is the screened count for that case.
This contrast evaluates point selection at a fixed count, but the rule does not include the mask-only incremental checks of the participant workflow.}

\subsection{\reviewrev{Participant screening and proposal order}}
\label{sec:current-policy}
\reviewrev{The participant workflow uses the candidate construction in main-paper Eq.~\ref{eq:prompt-location}.
Eligible components are prioritized using diagnostic score and component size.
\supprev{The policy suppresses recorded candidate identifiers and points less than 12\,mm from a same-label point in the feedback or refinement history.}
All remaining representatives enter review, without a top-$k$ cutoff.}

\reviewrev{For each point, the policy checks local response and whole-mask plausibility against the current mask, then checks incremental additions or removals against the mask-only response and considers supported-foreground loss.
Incremental directional support below 0.90 for positive points or 0.98 for negative points, and supported-foreground loss above 0.005, flag uncertain responses for user review.
A positive-point response is also checked for spatial separation from the current foreground.
The automatic verdict is \texttt{accept} when the required checks pass, \texttt{reject} for clear inconsistencies in the model response, or \texttt{review} when evidence remains uncertain.
These verdicts concern the output obtained with a single point; the user still chooses submitted points and approves or rejects their joint refinement.}

\section{Additional Qualitative Case Studies}
\label{sec:additional-cases}
\rev{Figure~\ref{fig:additional-cases} extends the KiTS19 and Pancreas-CT examples
in main-paper Fig.~\ref{fig:case-study} with cases from AortaSeg24, MSD Liver,
MSD Pancreas, MSD Spleen, and SLIVER07. \reviewrev{The panel follows rough masks through diagnosis and interaction to final masks and GT.
Selected slices illustrate the workflow, not success frequency or whole-volume correctness.}}

%% file: main.bbl
\begin{thebibliography}{29}
\providecommand{\natexlab}[1]{#1}
\providecommand{\url}[1]{\texttt{#1}}
\expandafter\ifx\csname urlstyle\endcsname\relax
  \providecommand{\doi}[1]{doi: #1}\else
  \providecommand{\doi}{doi: \begingroup \urlstyle{rm}\Url}\fi

\bibitem[{\c{C}}i{\c{c}}ek et~al.(2016){\c{C}}i{\c{c}}ek, Abdulkadir, Lienkamp, Brox, and Ronneberger]{Cicek2016}
{\"O}zg{\"u}n {\c{C}}i{\c{c}}ek, Ahmed Abdulkadir, Soeren~S. Lienkamp, Thomas Brox, and Olaf Ronneberger.
\newblock {3D} {U-Net}: Learning dense volumetric segmentation from sparse annotation.
\newblock In \emph{Proceedings of International Conference on Medical Image Computing and Computer Assisted Interventions}, pages 424--432, 2016.

\bibitem[Diaz-Pinto et~al.(2024)Diaz-Pinto, Alle, Nath, Tang, Ihsani, Asad, et~al.]{DiazPinto2022}
Andres Diaz-Pinto, Sachidanand Alle, Vishwesh Nath, Yucheng Tang, Alvin Ihsani, Muhammad Asad, et~al.
\newblock {MONAI Label}: A framework for {AI}-assisted interactive labeling of {3D} medical images.
\newblock \emph{Medical Image Analysis}, 95:\penalty0 103207, 2024.

\bibitem[Du et~al.(2023)Du, Bai, Huang, and Zhao]{du2023segvol}
Yuxin Du, Fan Bai, Tiejun Huang, and Bo Zhao.
\newblock {SegVol}: Universal and interactive volumetric medical image segmentation.
\newblock \emph{arXiv preprint arXiv:2311.13385}, 2023.

\bibitem[Fuchs et~al.(2022)Fuchs, Gonz{\'a}lez, and Mukhopadhyay]{Fuchs2022}
Moritz Fuchs, Camila Gonz{\'a}lez, and Anirban Mukhopadhyay.
\newblock Practical uncertainty quantification for brain tumor segmentation.
\newblock In \emph{Proceedings of International Conference on Medical Imaging with Deep Learning}, pages 407--422, 2022.

\bibitem[Hatamizadeh et~al.(2022{\natexlab{a}})Hatamizadeh, Nath, Tang, Yang, Roth, and Xu]{hatamizadeh2022swin}
Ali Hatamizadeh, Vishwesh Nath, Yucheng Tang, Dong Yang, Holger~R Roth, and Daguang Xu.
\newblock {Swin UNETR}: Swin {Transformers} for semantic segmentation of brain tumors in {MRI} images.
\newblock In \emph{Proceedings of the International Conference on Medical Image Computing and Computer Assisted Interventions Brainlesion Workshop}, pages 272--284, 2022{\natexlab{a}}.

\bibitem[Hatamizadeh et~al.(2022{\natexlab{b}})Hatamizadeh, Tang, Nath, Yang, Myronenko, Landman, Roth, and Xu]{hatamizadeh2022unetr}
Ali Hatamizadeh, Yucheng Tang, Vishwesh Nath, Dong Yang, Andriy Myronenko, Bennett Landman, Holger~R Roth, and Daguang Xu.
\newblock {UNETR}: Transformers for {3D} medical image segmentation.
\newblock In \emph{Proceedings of the IEEE/CVF Winter Conference on Applications of Computer Vision}, pages 1748--1758, 2022{\natexlab{b}}.

\bibitem[He et~al.(2024)He, Guo, Tang, Myronenko, Nath, Xu, Yang, Zhao, Simon, Belue, Harmon, Turkbey, Xu, and Li]{he2024vista3d}
Yufan He, Pengfei Guo, Yucheng Tang, Andriy Myronenko, Vishwesh Nath, Ziyue Xu, Dong Yang, Can Zhao, Benjamin Simon, Mason Belue, Stephanie Harmon, Baris Turkbey, Daguang Xu, and Wenqi Li.
\newblock {VISTA3D}: A unified segmentation foundation model for {3D} medical imaging.
\newblock \emph{arXiv preprint arXiv:2406.05285}, 2024.

\bibitem[Heller et~al.(2021)Heller, Isensee, Maier-Hein, Hou, Xie, Li, Nan, Mu, Lin, Han, et~al.]{Kits19}
Nicholas Heller, Fabian Isensee, Klaus~H. Maier-Hein, Xiaoshuai Hou, Chunmei Xie, Fengyi Li, Yang Nan, Guangrui Mu, Zhiyong Lin, Miofei Han, et~al.
\newblock The state of the art in kidney and kidney tumor segmentation in contrast-enhanced {CT} imaging: Results of the {KiTS19} challenge.
\newblock \emph{Medical Image Analysis}, 67:\penalty0 101821, 2021.

\bibitem[Imran et~al.(2026)Imran, Krebs, Sivaraman, Zhang, Kumar, Ueland, Fassler, Wang, Huang, Sun, et~al.]{AortaSeg24}
Muhammad Imran, Jonathan~R. Krebs, Vishal~Balaji Sivaraman, Teng Zhang, Amarjeet Kumar, Walker~R. Ueland, Michael~J. Fassler, Chang Wang, Jinlong Huang, Xiao Sun, et~al.
\newblock Multi-class segmentation of aortic branches and zones in computed tomography angiography: The {AortaSeg24} challenge.
\newblock \emph{Medical Image Analysis}, 113:\penalty0 104188, 2026.

\bibitem[Isensee et~al.(2021)Isensee, Jaeger, Kohl, Petersen, and Maier-Hein]{isensee2021nnu}
Fabian Isensee, Paul~F. Jaeger, Simon A.~A. Kohl, Jens Petersen, and Klaus~H. Maier-Hein.
\newblock {nnU-Net}: A self-configuring method for deep learning-based biomedical image segmentation.
\newblock \emph{Nature Methods}, 18\penalty0 (2):\penalty0 203--211, 2021.

\bibitem[Isensee et~al.(2025)Isensee, Rokuss, Kr{\"a}mer, Dinkelacker, Ravindran, Stritzke, Hamm, Wald, Langenberg, Ulrich, et~al.]{isensee2025nninteractive}
Fabian Isensee, Maximilian Rokuss, Lars Kr{\"a}mer, Stefan Dinkelacker, Ashis Ravindran, Florian Stritzke, Benjamin Hamm, Tassilo Wald, Moritz Langenberg, Constantin Ulrich, et~al.
\newblock {nnInteractive}: Redefining {3D} promptable segmentation.
\newblock \emph{arXiv preprint arXiv:2503.08373}, 2025.

\bibitem[Kirillov et~al.(2023)Kirillov, Mintun, Ravi, Mao, Rolland, Gustafson, Xiao, Whitehead, Berg, Lo, Doll{\'a}r, and Girshick]{SAM}
Alexander Kirillov, Eric Mintun, Nikhila Ravi, Hanzi Mao, Chloe Rolland, Laura Gustafson, Tete Xiao, Spencer Whitehead, Alexander~C. Berg, Wan-Yen Lo, Piotr Doll{\'a}r, and Ross Girshick.
\newblock Segment anything.
\newblock In \emph{Proceedings of IEEE/CVF International Conference on Computer Vision}, pages 4015--4026, 2023.

\bibitem[Liu et~al.(2026)Liu, Bao, Yang, Geng, Zheng, Li, Chen, Peng, and Yuan]{Liu2026}
Shengyuan Liu, Liuxin Bao, Qi Yang, Wanting Geng, Boyun Zheng, Chenxin Li, Wenting Chen, Houwen Peng, and Yixuan Yuan.
\newblock {MedSAM-Agent}: Empowering interactive medical image segmentation with multi-turn agentic reinforcement learning.
\newblock \emph{arXiv preprint arXiv:2602.03320}, 2026.

\bibitem[Ma et~al.(2024)Ma, He, Li, Han, You, and Wang]{MedSAM}
Jun Ma, Yuting He, Feifei Li, Lin Han, Chenyu You, and Bo Wang.
\newblock Segment anything in medical images.
\newblock \emph{Nature Communications}, 15\penalty0 (1):\penalty0 654, 2024.

\bibitem[Milletari et~al.(2016)Milletari, Navab, and Ahmadi]{milletari2016v}
Fausto Milletari, Nassir Navab, and Seyed-Ahmad Ahmadi.
\newblock {V-Net}: Fully convolutional neural networks for volumetric medical image segmentation.
\newblock In \emph{Proceedings of the Fourth International Conference on 3D Vision}, pages 565--571, 2016.

\bibitem[Reen et~al.(2026)Reen, Borkar, and Mahapatra]{Reen2026}
Abhilaksh~Singh Reen, Kushal Borkar, and Ritvik Mahapatra.
\newblock {IMVS}: Interactive medical volume segmentation with test-time adaptation: A new method for annotating radiology datasets.
\newblock \emph{arXiv preprint arXiv:2609.16775}, 2026.

\bibitem[Ronneberger et~al.(2015)Ronneberger, Fischer, and Brox]{Ronneberger2015}
Olaf Ronneberger, Philipp Fischer, and Thomas Brox.
\newblock {U-Net}: Convolutional networks for biomedical image segmentation.
\newblock In \emph{Proceedings of International Conference on Medical Image Computing and Computer Assisted Interventions}, pages 234--241, 2015.

\bibitem[Roth et~al.(2016)Roth, Farag, Turkbey, Lu, Liu, and Summers]{pancreasct}
Holger Roth, Amal Farag, Evrim~B. Turkbey, Le Lu, Jiamin Liu, and Ronald~M. Summers.
\newblock Data from pancreas-{CT}, 2016.

\bibitem[Shaker et~al.(2024)Shaker, Maaz, Rasheed, Khan, Yang, and Khan]{shaker2022unetr++}
Abdelrahman Shaker, Muhammad Maaz, Hanoona Rasheed, Salman Khan, Ming-Hsuan Yang, and Fahad~Shahbaz Khan.
\newblock {UNETR++}: Delving into efficient and accurate {3D} medical image segmentation.
\newblock \emph{IEEE Transactions on Medical Imaging}, 43\penalty0 (9):\penalty0 3377--3390, 2024.

\bibitem[Simpson et~al.(2019)Simpson, Antonelli, Bakas, et~al.]{simpson2019large}
Amber~L Simpson, Michela Antonelli, Spyridon Bakas, et~al.
\newblock A large annotated medical image dataset for the development and evaluation of segmentation algorithms.
\newblock \emph{arXiv preprint arXiv:1902.09063}, 2019.

\bibitem[Specktor{-}Fadida et~al.(2025)Specktor{-}Fadida, Ben{-}Sira, Ben{-}Bashat, and Joskowicz]{SpecktorFadida2025}
Bella Specktor{-}Fadida, Liat Ben{-}Sira, Dafna Ben{-}Bashat, and Leo Joskowicz.
\newblock {SegQC}: A segmentation network-based framework for multi-metric segmentation quality control and segmentation error detection in volumetric medical images.
\newblock \emph{Medical Image Analysis}, 103:\penalty0 103638, 2025.

\bibitem[Van~Ginneken et~al.(2007)Van~Ginneken, Heimann, and Styner]{van07}
Bram Van~Ginneken, Tobias Heimann, and Martin Styner.
\newblock {3D} segmentation in the clinic: A grand challenge.
\newblock In \emph{Proceedings of the International Conference on Medical Image Computing and Computer-assisted Intervention Workshop on 3D Segmentation in the Clinic: A Grand Challenge}, pages 7--15, 2007.

\bibitem[Wang et~al.(2018)Wang, Li, Zuluaga, Pratt, Patel, Aertsen, Doel, David, Deprest, Ourselin, and Vercauteren]{wang2018interactive}
Guotai Wang, Wenqi Li, Maria~A. Zuluaga, Rosalind Pratt, Premal~A. Patel, Michael Aertsen, Tom Doel, Anna~L. David, Jan Deprest, S{\'e}bastien Ourselin, and Tom Vercauteren.
\newblock Interactive medical image segmentation using deep learning with image-specific fine tuning.
\newblock \emph{IEEE Transactions on Medical Imaging}, 37\penalty0 (7):\penalty0 1562--1573, 2018.

\bibitem[Wasserthal et~al.(2023)Wasserthal, Breit, Meyer, Pradella, Hinck, Sauter, Heye, Boll, Cyriac, Yang, Bach, and Segeroth]{wasserthal2023totalsegmentator}
Jakob Wasserthal, Hanns-Christian Breit, Manfred~T. Meyer, Maurice Pradella, Daniel Hinck, Alexander~W. Sauter, Tobias Heye, Daniel Boll, Joshy Cyriac, Shan Yang, Michael Bach, and Martin Segeroth.
\newblock {TotalSegmentator}: Robust segmentation of 104 anatomic structures in {CT} images.
\newblock \emph{Radiology: Artificial Intelligence}, 5\penalty0 (5):\penalty0 e230024, 2023.

\bibitem[Wong et~al.(2024)Wong, Rakic, Guttag, and Dalca]{wong2024scribbleprompt}
Hallee~E Wong, Marianne Rakic, John Guttag, and Adrian~V Dalca.
\newblock {ScribblePrompt}: Fast and flexible interactive segmentation for any biomedical image.
\newblock In \emph{Proceedings of the European Conference on Computer Vision}, pages 207--229, 2024.

\bibitem[Zaman et~al.(2023)Zaman, Zhang, Zhang, Sonka, and Wu]{Zaman2023}
Fahim~A. Zaman, Lichun Zhang, Honghai Zhang, Milan Sonka, and Xiaodong Wu.
\newblock Segmentation quality assessment by automated detection of erroneous surface regions in medical images.
\newblock \emph{Computers in Biology and Medicine}, 164:\penalty0 107324, 2023.

\bibitem[Zhang et~al.(2021)Zhang, Shi, Sun, Wang, Zhou, Gao, and Shen]{zhang2021interactive}
Jian Zhang, Yinghuan Shi, Jinquan Sun, Lei Wang, Luping Zhou, Yang Gao, and Dinggang Shen.
\newblock Interactive medical image segmentation via a point-based interaction.
\newblock \emph{Artificial Intelligence in Medicine}, 111:\penalty0 101998, 2021.

\bibitem[Zhao et~al.(2023)Zhao, Zhang, Wu, Zhang, Zhang, Wang, and Xie]{zhao2023sat}
Ziheng Zhao, Yao Zhang, Chaoyi Wu, Xiaoman Zhang, Ya Zhang, Yanfeng Wang, and Weidi Xie.
\newblock One model to rule them all: Towards universal segmentation for medical images with text prompts.
\newblock \emph{arXiv preprint arXiv:2312.17183}, 2023.

\bibitem[Zhou et~al.(2021)Zhou, Guo, Zhang, Yu, Wang, and Yu]{zhou2021nnformer}
Hong-Yu Zhou, Jiansen Guo, Yinghao Zhang, Lequan Yu, Liansheng Wang, and Yizhou Yu.
\newblock {nnFormer}: Interleaved {Transformer} for volumetric segmentation.
\newblock \emph{arXiv preprint arXiv:2109.03201}, 2021.

\end{thebibliography}
